# Event-based YOLO Object Detection: Proof of Concept for Forward Perception System

Waseem Shariff*[1,2], Muhammad Ali Farooq[1], Joe Lemley[2], and Peter Corcoran[1]
[1]University of Galway
[2]Xperi Corporation

## ABSTRACT

Neuromorphic vision or event vision is an advanced vision technology, where in contrast to visible camera sensors that output pixels, the event vision generates neuromorphic events every time there's a brightness change which exceeds a specific threshold in the field of view (FoV). This study focuses on leveraging neuromorphic event data for roadside object detection. This is a proof of concept towards building artificial intelligence (AI) based pipelines which can be used for forward perception systems for advanced vehicular applications. The focus is on building efficient state-of-the-art object detection networks with better inference results for fast-moving forward perception using an event camera. In this article, the event simulated A2D2 dataset is manually annotated and trained on two different YOLOv5 networks (small and large variants). To further assess its robustness, single model testing and ensemble model testing are carried out.

## 1. INTRODUCTION

Low latency, a very wide dynamic range, and low power consumption are just a few of the exciting features that event cameras provide for an integrated vehicle sensor suite. These sensors are distinctive in their own way. Event cameras react to brightness changes in the field of vision both asynchronously and independently. With each change in brightness, the event camera triggers a sequence of variable data called "*Spike*" or "*Event*" [1]. Every time a sensor outputs such an event, it learns the logged brightness and constantly checks for a change from its stored brightness value. If it exceeds a particular threshold, then the camera transmits an event, with the $x, y$ position, the timestamp $t$, as well as the 1-bit polarity of the changes whether it was a positive change in brightness (1) or negative change (0/-1).

The output of event camera sensors is data-driven, which is based on how much the scene is in motion or changing in brightness [1]. This is a big advantage to the embedded forward perception automotive applications since there is constant change in the vehicle's field of view. Vehicle perception systems are becoming an evolving consumer technology application, and as they advance over time, they will potentially offer more safety advantages and dependable modes of transportation [2]. Moreover, event cameras are of great interest to researchers in the field of automotive. The research community is currently focusing on tackling event-based vision for vehicular sensor suites.

In this paper, we propose neuromorphic-event guided object detection (OD) using the most efficient and reliable OD framework -YOLOv5 [3]. YOLO algorithms use regression methods to learn an entire image at once while improving the performance globally. Additionally, it detects class objects simultaneously together with their probability scores without the need for region feedback, which makes it the most effective method for detecting objects [4]. Events in this study are converted to 2d frames since YOLOv5 uses 2d frames as inputs. Event frames are the data structure appropriate with traditional computer vision since they contain spatial information about scene boundaries and provide information about events' absence as well as their existence, which is useful [1] in object detection. Object detection plays a vital role in vehicle perception systems. Nevertheless, in the field of computer vision, the adaptive capacity of object detection algorithms using dense event data remains a difficult research field with varied outcomes [5]. The key cause for this is, most of the current research use low-resolution event data to represent sparse-based object detection and there are no recent studies on high-resolution event-based object detection. And it is also challenging to converge deep learning models on high resolution event data features. To overcome such flaws and normalize use of synthetic data, this study uses the A2D2 public dataset [6], a visible frame-based dataset. Since the dataset comprises good *continuous frames* with a high resolution of 640x480, this helps synthesize visible to event frames easy. Continuous frames assist in the better synthesis of visible frames to event frames as the event simulator can detect a distinct brightness difference between two frames.

The main goal is to modify the state-of-the-art (SoA) object detection framework for an event-based vision for four common classes which include persons, cars, poles, and other vehicles. The network is trained on 3146 synthetics and augmented event samples, and further, the network is evaluated on 546 synthetic event frames. The performance of the trained event network is measured using various quantitative metrics which include precision, recall, mean average precision and frames per second. In this paper, the major contribution lies in training two different network variants of YOLOv5 which includes a small and large network. During the training phase, we opted for network hyperparameters along with an SGD optimizer for optimal tuning of CNN-based networks on event data. Moreover, two separate test methodologies, single model testing (SMT) and ensemble model testing (EMT) are used in the overall validation testing to boost test accuracy.

## 2. BACKGROUND

Recently, event-based object detection has become a popular subject of interest. Different approaches to representing event data and performing object classification are being developed by researchers. These techniques employ state-of-the-art (SoA) spike neural networks [7], histograms [8] and many more [9][10][11], which take advantage of the data's sparsity and highlight the temporal nature of the event camera. These techniques are validated successfully, nevertheless, on low-resolution classification tasks. Moreover, event inputs with substantially high event rates and larger resolution are still a challenge [4]. On the other hand, event dense representation can be used with SoA computer vision architectures. Although some of the best approaches that were taken are event-RGB fusion for object detection [12][13], remote monitoring-object recognition applications [14], transfer-learning [15], and nearest neighbors [16], these techniques were only evaluated on simplified action scenes that feature a few moving objects against a still background. Further authors in [17], proposed a mixed event (dynamic vision sensor)-frame (active pixel sensor) based pedestrian detection using convolutional neural networks (CNN). This mixed approach was evaluated with the best average precision of 92% using YOLOv3. Jixiang Wan et al [18], proposed a novel event-to-frame-based approach further to detect pedestrian scenes. The network performed with an average precision of 81.43% at 26 frames per second (FPS).

This study focuses on enhancing the performance and inference time of feed-forward techniques. Moreover, this study also focuses on major roadside objects which include cars, pedestrians, poles, and other vehicles. Furthermore, the scarcity of benchmarked event-based high-resolution datasets for object detection is also a factor. In this study, the A2D2 dataset [6] with a high resolution of 640x480 is simulated to event frames using v2e [19] to train YOLOv5 networks.

## 3. PROPOSED METHODOLOGY

In this section, the proposed methodology and the approach for effective training of the state-of-the-art YOLO-v5 framework for out-of-cabin object detection in the neuromorphic event domain are shown. YOLO algorithm utilizes a regression approach to train the entire image at once to improve the overall performance of the model. Additionally, it recognizes class objects simultaneously together with their respective probability scores. In this paper, the networks are trained to detect four distinct moving and stationery objects which include pedestrians/people, cars, other vehicles (such as buses, vans, or trucks), and street light poles. These objects are commonly found on the roadside, and it helps for better development of the driver's assistant system. Figure 1 shows the block diagram of the proposed methodology for Event-based YOLO object detection.

### 3.1 Data Transformation

Event cameras are extremely sensitive to intensity changes in the field of view, so we can observe a lot of noise in real event datasets. Since there might be thousands of events generated every second, it is still a research topic for the scientific community to determine which event is good and how to sub-sample the events. In this study, using an approach that is based on V2E [19], we simulate successive sequences of visible frames to event frames. Simulating event frames has the benefit of allowing simulators to identify and highlight distinct differences between two frames thus saving us from performing sub-sampling operations. To log the intensity value for each frame, v2e first transforms all frames into M-Luma frames with respective timestamps. Additionally, v2e specifies the temporal and logarithmic contrast thresholds for each frame to evaluate the intensity variation between frames. Finally, it generates events depending on the temporal difference and threshold mismatch [19]. As can be observed from Figure 1, the data transformation block shows a sample of event generation from the visible frame. With this preprocessing step, the intensity frames are converted to synthetic event frames with realistic temporal and leak noise. Furthermore, 2281 frames from the A2D2 visible dataset [6] were extracted and simulated based on the visibility of most of the objects. The dataset was divided in the ratio of 75:25 before data augmentation, where 75% of the data was used in the training & validation process and the rest of the 25% was used for testing & evaluation purposes. Figure 2 shows the samples of the event objects from each class.

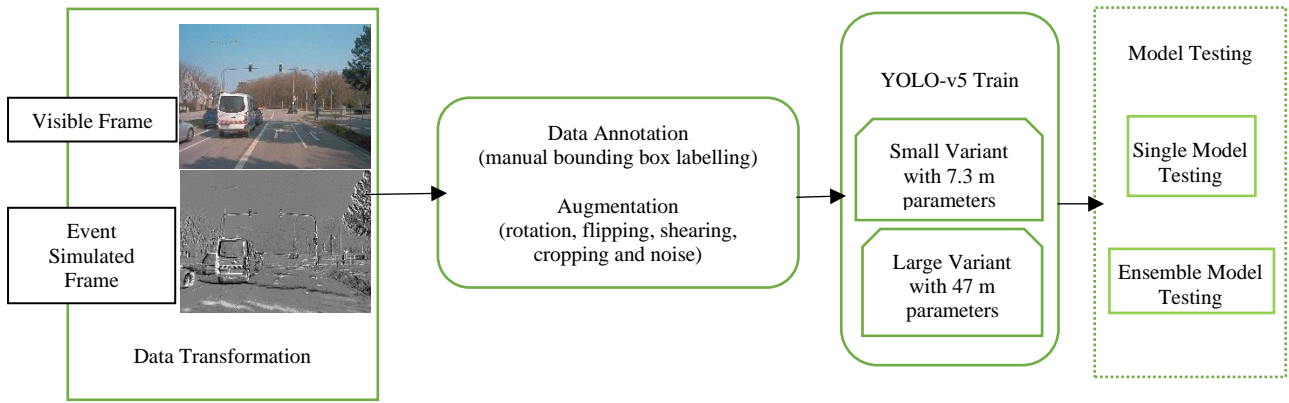

Figure 1. Event-based YOLO object detection.

## 3.2 Data Annotation and Augmentation

In this study, we have performed manual bounding box-based annotations for all the event classes. Both the Small and Large variant of YOLO was trained to detect person, pole, cars, and other vehicles. Further to overcome the limitations of the smaller dataset and perform optimal training for the networks, data augmentation and transformation techniques were applied to training data. The augmentation includes flipping, rotation, image cropping, and shearing of the data. With the augmentation techniques, the training data was increased by almost 2.5 times. Figure 3 shows the manual annotation/labelling done on some of the event data. The manual annotation and the augmentation that was performed on the samples are shown in Figure 3.

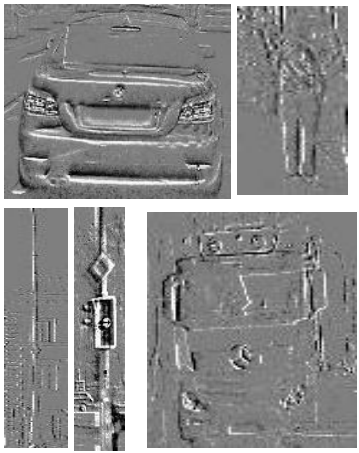

Figure 2. Four different classes (car, person, poles and other vehicles-truck, bus or van) are used for training the event object detection network.

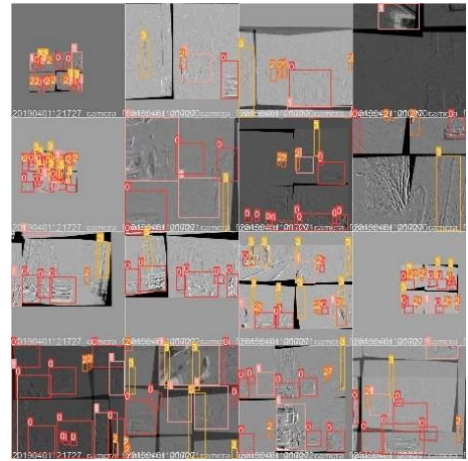

Figure 3. Different types of Augmentation techniques are applied.

## 3.3 Event-based YOLO training

As discussed in this study we have used a trade-off between two different models of YOLOv5. The first approach uses a small network variant to achieve a lower inference time. Whereas in the second approach we have used a larger network variant to achieve robust detection results, however, with slightly greater inference time. The complete training and testing process was carried out on a server machine with dedicated 12 GB GEFORCE RTX 2080 graphics processing units. The training was finetuned with a learning rate of 0.015 with 50 epochs. Utilizing a variety of quantitative metrics, the training accuracy and loss outcomes are examined to completely determine the effectiveness of all trained models. The YOLO-v5 framework estimates the total loss based on three different scores, including the bounding box regression score, the class probability score, and the object score (shown in figure 4). In addition to this, the model accuracy is calculated using the recall rate, precision, and mean average precision (mAP). Table 1 and Figure 5 show the mean average precision obtained from training both the small and large networks.

Table 1. The overall training performance of YOLOv5 small and large variant networks.

| YOLO-variant | Box Loss | Object Loss | Classification Loss | Precision | Recall | Mean Avg Precision (mAP) | Training Time |
|---|---|---|---|---|---|---|---|
| Small | 0.02232 | 0.01229 | 0.00175 | 77.5% | 75.3% | 77% | **34m 48s** |
| Large | **0.01708** | **0.00961** | **0.00103** | **79.2%** | **79.5%** | **82%** | 1h 53m 20s |

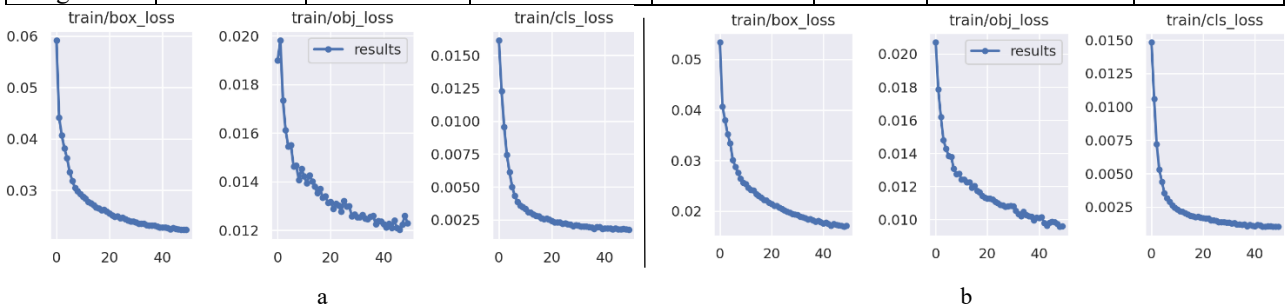

Figure 4. Training and Loss estimates of YOLOv5 networks with three different score box loss, object loss and classification loss (a) Small Variant (b) Large Variant

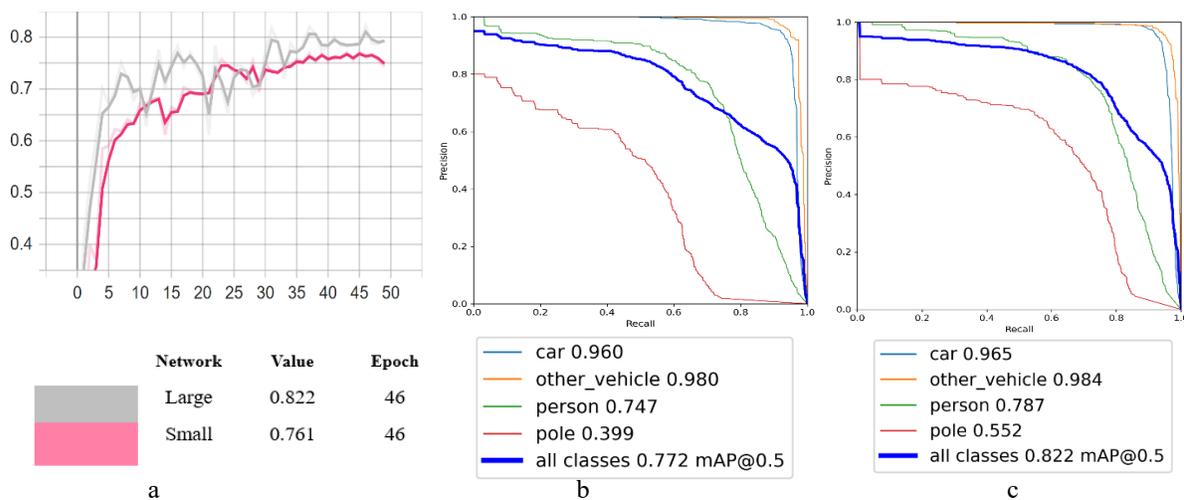

Figure 5. Training performances of both small and large variants of YOLOv5. (a) Mean Average Precision graph describing both small and large network training, (b) Precision-Recall Curve of Small network variant, (c) Precision-Recall Curve of Large network variant

For better understanding, table 1 and figures 4 & 5 show the numerical results of all accuracy metrics and loss metrics. It can further be summarized among the small and large variant networks; the large variant network training was better than the small variant and there are significant differences in the class-wise accuracy distributions. The best mean average precision of the large variant was 82.2% and 77.2% for the small variant.

## 4. TESTING RESULTS

As mentioned before in this study, there are two different approaches used which include single model testing (SMT) and ensemble model testing (EMT). SMT is referred to as a traditional testing strategy, where the unseen data is given to the trained object detection model. With this approach, the inference time is dependent on the single trained model. On contrary, EMT also known as ensemble learning in machine learning refers to the simultaneous use of two or more trained networks to create a single, ideal predictive inference model. Figure 6 shows the difference between the testing approaches of SMT and EMT.

In this study, two separately trained YOLO networks (small and large) were evaluated by running them in together to help in getting more better results by having improved mean-average precision (mAP). However, running multiple trained models together can increase the inference time per image which in turn results in decreased FPS. Thus, it is a tradeoff between accuracy and inference speed. The test dataset, consisting of 587 frames was used for the evaluation of the trained small and large models. For evaluating purposes, the IOU and confidence threshold was set to 0.2. The small network when evaluated using the SMT method, produced a mean average accuracy of 66% and best inference of 201 frames per second (FPS). A mean average accuracy of 67% was obtained for the large network variant using the SMT method with an inference speed of 62 FPS. Additionally, the EMT had the best mean average precision of 69% with an average inference of 45 frames per second. Figure 7 shows some of the test-inference results from the evaluation set.

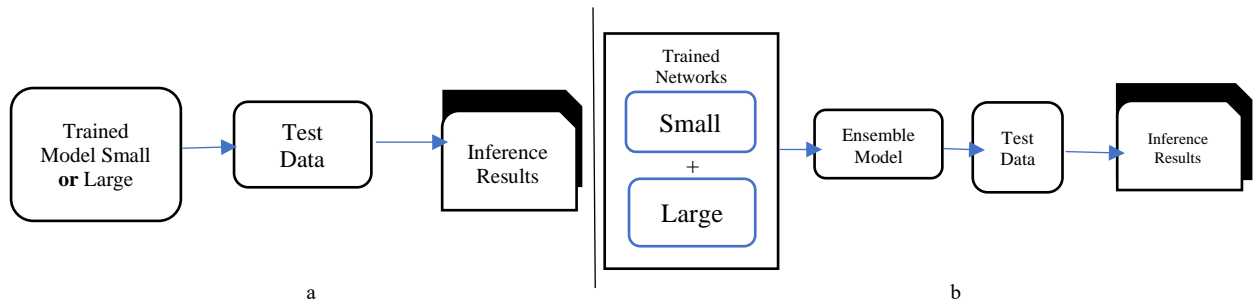

Figure 6. Evaluation Methodologies used are (a) Single Model Testing Illustrations, (b) Ensemble Model Testing Illustrations

Table 2. The overall testing performance of YOLOv5 small and large variant networks (using both SMT and EMT).

| Network Variant | Single Model Testing (SMT) | | Ensemble Model Testing (EMT) | |
|---|---|---|---|---|
| | Mean Average Precision (mAP) | Average Inference Time-FPS | Mean Average Precision (mAP) | Average Inference Time-FPS |
| Small | 66% | **201** | | |
| Large | **67%** | 62 | **69%** | 45 |

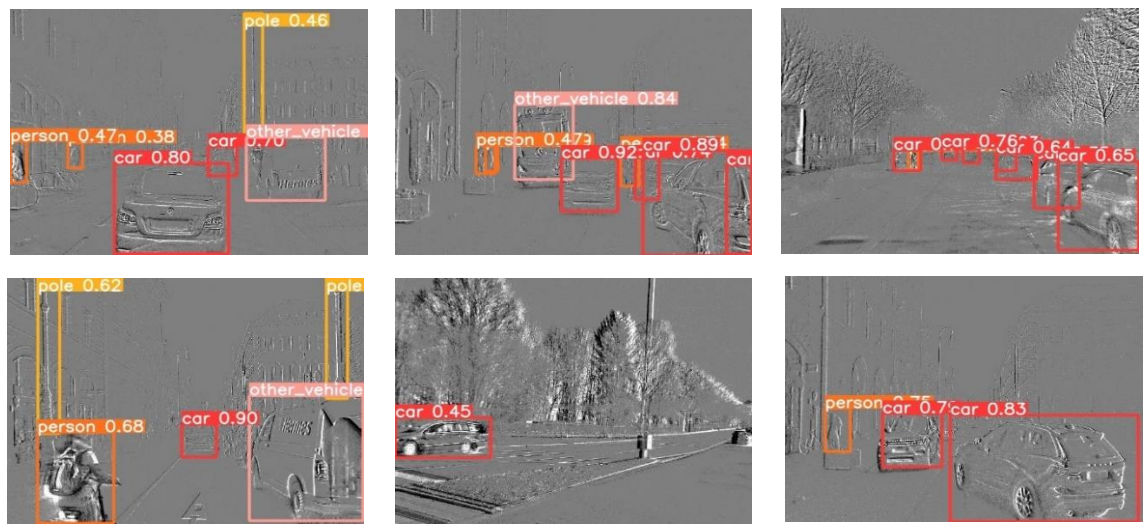

Figure 7. Testing results- object detection on event frames with corresponding. These results were extracted from SMT-testing and EMT-testing

The following are some of the samples of the extreme complex frames found during the experimental analysis. Figure 8 demonstrates the trained model's robustness. The trained model was able to locate the objects despite their great distance. Secondly, the trained detector was able to detect the objects even though the car had stopped and the brightness in the field of vision had barely changed. Figure 9 shows some of the misclassifications of the trained model. In the first frame, a store window on the left side along the roadside was misclassified as *other_vehicle*. In the second frame, it can be seen that a plant obstacle on the left was incorrectly classified as a *car*, and finally, in the third frame, on-road markings were mistakenly identified as *poles*.

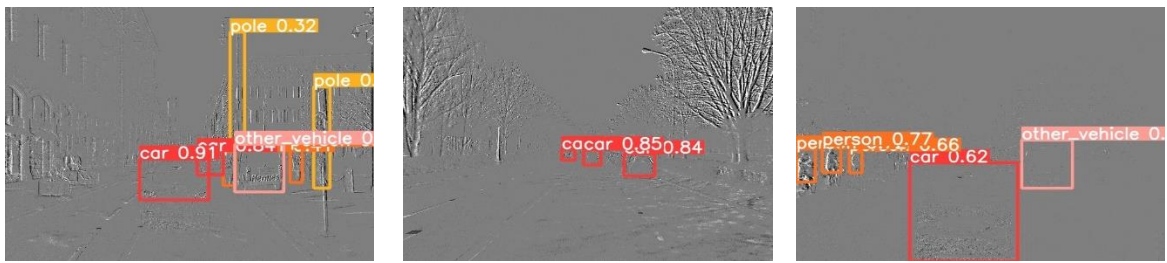

Figure 8. Few examples of severe scenes that show the robust nature of trained models.

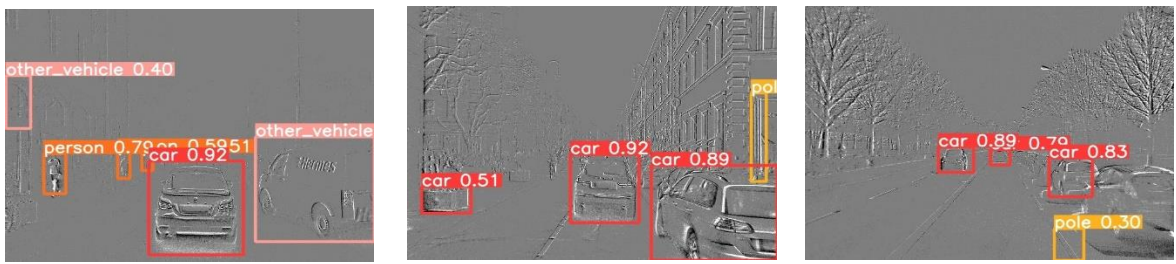

Figure 9. Scenario instances illustrate misclassifications.

## 5. CONCLUSION

In this article, we have proposed a proof of concept for a smart event-based forward perception system. The simulated event data was manually annotated which includes common roadside objects which include persons, cars, poles, and other vehicles (like trucks, buses, or vans). For this study, we have focused on training two YOLOv5 networks which incorporate small variants and large variants. The large variant achieved a maximum mean average precision (mAP) accuracy of 82% and the small variant achieved 77% mAP in the training phase. The performance estimation was validated using two different methods, which include SMT-single model testing and EMT-ensemble model testing. During the SMT, the small network variant achieved a maximum mean average precision accuracy of 66% and the large network variant achieved 67%. Further the small network had the highest FPS of 201 and the large network achieved 62 FPS. On the contrary, EMT achieved better mAP accuracy of 69% but with only 47 FPS. The most important finding from the overall performance observation is that, although achieving comparable accuracy to the large variant network and ensemble-based testing, the small network has the fastest inference of 201 frames per second.

In the future, the dataset will contain more samples to overcome the misclassification rate. To make the dataset more diverse and durable, it will also include both simulated events and real-time event data. Additionally, the FPS can be further improved by utilizing inference accelerators, which will increase their efficiency and reduce their computational cost for real-time deployment on embedded low-power devices like the Nvidia Jetson Nano and Jetson Xavier.

## ACKNOWLEDGMENTS

This is study is funded by FotoNation, Ltd. (a part of Xperi Corporation), under the Irish Research Council Employment PhD Program.

## AUTHORS' BACKGROUND

| Your Name | Title* | Research Field | Personal website |
| --- | --- | --- | --- |
| Waseem Shariff | PhD Candidate | Neuromorphic event vision for vehicular | https://www.linkedin.com/in/waseem-shariff-997534141/ |

| | | perception and in-cabin driver monitoring system. | |
|---|---|---|---|
| Dr Muhammad Ali Farooq | Postdoc Researcher | Thermal Augmented Awareness, an extended perception for enabling safe autonomous driving | https://www.linkedin.com/in/muhammad-ali-farooq-phd-876235a1/ |
| Dr Joe Lemley | R&D Principal Engineer, Xperi | Automotive driver monitoring and biometrics | https://www.linkedin.com/in/joe-lemley-0b442245/ |
| Dr Peter Corcoran | Professor | Recognized for his contribution to digital camera technology notably in-camera red-eye correction and facial detection. Currently focused on state-of-the-art computer-vision imaging and Edge-AI. | https://www.universityofgalway.ie/our-research/people/engineering-and-informatics/petercorcoran/ |

*This form helps us to understand your paper better, the form itself will not be published.

*Title can be chosen from: master student, PhD candidate, assistant professor, lecture, senior lecture, associate professor, full professor